\documentclass[12pt]{article}

\usepackage[margin=1in]{geometry}
\usepackage{setspace}
\usepackage{listings}
\usepackage[toc,page,header]{appendix}
\usepackage{etoc}
\usepackage{minitoc}
\usepackage{ragged2e}
\usepackage[none]{hyphenat}
\usepackage{verbatim}
\usepackage[utf8]{inputenc}
\usepackage[T1]{fontenc}
\usepackage[sc]{mathpazo}
\usepackage{microtype}

\usepackage{natbib}

\usepackage{amsmath}
\usepackage{siunitx}

\usepackage{graphicx}
\usepackage{placeins}
\usepackage{lscape}
\usepackage{bbm}
\usepackage{tabularx, booktabs}
\usepackage{multirow}
\usepackage{longtable}
\usepackage{float}
\usepackage{subcaption}
\usepackage{caption}
\usepackage[most]{tcolorbox}
\usepackage{listings}

\lstdefinestyle{mystyle}{
    basicstyle=\ttfamily\footnotesize,
    columns=fullflexible,
    breaklines=true,    
    postbreak=\mbox{\textcolor{red}{$\hookrightarrow$}\space},
    captionpos=b,                    
    keepspaces=true,                 
    numbers=left,                    
    numbersep=2pt,                  
    showspaces=false,                
    showstringspaces=false,
    showtabs=false,                  
    tabsize=1
}
\usepackage{url}  
\usepackage[super]{nth}

\usepackage{hyperref}
\hypersetup{
	colorlinks=true,
    urlcolor=blue,
	linkcolor=black,
	citecolor=black
}

\title{Agent-Enhanced Large Language Models\\for Researching Political Institutions\thefootnote\relax\footnotetext{The authors thank attendees of the AI and the Study of Political Institutions Conference at the University of Southern California's Political Institutions and Political Economy (PIPE) Collaborative for helpful comments. Example code for \texttt{CongressRA}, the proof-of-concept LLM agent highlighted in this paper, can be found at \href{https://github.com/congressRA/sample-agent}{\tt https://github.com/congressRA/sample-agent}.}}
\author{Joseph R. Loffredo\thanks{PhD Candidate, Department of Political Science, MIT. \href{mailto:loffredo@mit.edu}{\tt loffredo@mit.edu}} \and Suyeol Yun\thanks{Independent Researcher. \href{mailto:syyun@alum.mit.edu}{\tt syyun@alum.mit.edu}}}
\date{March 15, 2025}

\begin{document}

\maketitle
\thispagestyle{empty}
\begin{abstract}
    \noindent The applications of Large Language Models (LLMs) in political science are rapidly expanding. This paper demonstrates how LLMs, when augmented with predefined functions and specialized tools, can serve as dynamic agents capable of streamlining tasks such as data collection, preprocessing, and analysis. Central to this approach is agentic retrieval-augmented generation (Agentic RAG), which equips LLMs with action-calling capabilities for interaction with external knowledge bases. Beyond information retrieval, LLM agents may incorporate modular tools for tasks like document summarization, transcript coding, qualitative variable classification, and statistical modeling. To demonstrate the potential of this approach, we introduce \texttt{CongressRA}, an LLM agent designed to support scholars studying the U.S. Congress. Through this example, we highlight how LLM agents can reduce the costs of replicating, testing, and extending empirical research using the domain-specific data that drives the study of political institutions.
\end{abstract}
\textbf{Keywords: large language models, retrieval-augmented generation, AI agent, Agentic RAG, political institutions} 
\vfill

\clearpage
\pagenumbering{arabic}
\doublespacing
\RaggedRight
\section*{Introduction}

The applications of large language models (LLM) in political science have grown significantly in the last few years, with recent review articles \citep[e.g.,][]{linegar_large_2023, ziems_can_2024, lee_applications_2024} highlighting their utility for a range of tasks, including content generation \citep{bail_can_2024}, measurement scale creation \citep{wu_large_2023}, simulating human behavior \citep{binz_using_2023, dillion2023can, grossmann_ai_2023, baker_simulating_2024, mei_turing_2024, xie_can_2024}, and text analysis \citep{gilardi_chatgpt_2023, nay_large_2023, halterman_codebook_2024, heseltine_large_2024, liu_poliprompt_2024, ornstein_how_2024, tornberg_large_2024}. In this paper, we argue and demonstrate that LLMs, when enhanced with predefined functions and equipped with specialized tools, can serve as agents \citep{xi_rise_2023}. By implementing LLM agents, researchers are capable of efficiently completing complex and time-intensive tasks like processing large datasets, constructing measures using multiple data sources, and preparing analyses.

While LLMs offer significant advantages, they also face well-documented limitations, such as generating incorrect, outdated, or irrelevant information and relying on non-credible sources \citep{ji_survey_2023, zhang_agentic_2024}. These issues, often referred to as "hallucinations," stem from their reliance on static, pre-trained knowledge bases that cannot be easily updated or verified without extensive retraining. To address these challenges, we propose that political science scholars implement and utilize LLMs with agentic retrieval-augmented generation (Agentic RAG) \citep{ravuru_agentic_2024, lewis_retrieval-augmented_2020}. By equipping LLMs with action-calling capabilities, Agentic RAG enables interaction with external knowledge bases---such as SQL databases, vector databases, or graph databases---as well as APIs and web-based resources. These LLM agents can autonomously determine the most relevant sources to query and help synthesize accurate, contextually appropriate responses \citep{an_golden-retriever_2024, gupta_comprehensive_2024}. This ability significantly reduces hallucinations and enhances reliability of LLM output.

Beyond dynamic retrieval capabilities, LLM agents can package additional functionalities into a single modular system. By integrating tools to perform actions such as document summarization, classification, and statistical analysis, researchers gain a flexible and transportable system. This means the same agent can be adapted to a research context by providing task-specific instructions, while leveraging a consistent set of shared tools that remain relevant across various studies. To demonstrate the potential of LLM agents in assisting with the study of political institutions, we introduce \texttt{CongressRA} as a proof of concept, an LLM agent designed to support scholars studying the U.S. Congress. We illustrate its utility by reconstructing the "legislative gridlock" index developed by \citet{binder_dynamics_1999}. To assist researchers, we have made sample code for developing and implementing a similar LLM agent.\footnote{Sample code for the proof-of-concept LLM agent (\texttt{CongressRA}) showcased in this paper is available at \href{https://github.com/congressRA/sample-agent}{\tt https://github.com/congressRA/sample-agent}.} Through this example, we highlight how the modularity and adaptability of agents like \texttt{CongressRA} can reduce the costs of replicating, testing, and extending empirical research in political science.

\section*{The Potential of LLM Agents}

Large language models (LLM) have garnered significant attention from those aiming to leverage these tools for Natural Language Processing (NLP). In particular, LLMs are highly effective in Natural Language Generation (NLG), producing coherent and contextually relevant text from structured or unstructured data \citep{gatt_survey_2018}. LLMs are neural network-based and typically built on the transformer architecture \citep{vaswani_attention_2023}. By adopting the transformer architecture, LLMs output vectors as contextualized word vector embeddings that capture the underlying or semantic meaning of inputted text. Embeddings can be used for tasks like finding pieces of information by comparing their numerically computable similarities, such as cosine similarity and \(L_p\) norms. Furthermore, LLMs show a basic form of human-like intelligence, allowing them to handle tasks while understanding the context and intent of a user's prompt.

Because of these capabilities, LLMs have a range of potential applications useful to scholars of political science \citep{linegar_large_2023, ziems_can_2024, lee_applications_2024}. These applications include content generation \citep{bail_can_2024}, measurement scale creation \citep{wu_large_2023}, simulating human behavior \citep{binz_using_2023, dillion2023can, grossmann_ai_2023, baker_simulating_2024, mei_turing_2024, xie_can_2024}, and text analysis \citep{gilardi_chatgpt_2023, nay_large_2023, halterman_codebook_2024, heseltine_large_2024, liu_poliprompt_2024, ornstein_how_2024, tornberg_large_2024}. The accessibility of these applications has grown due to the availability of open-source models and APIs, enabling scholars to utilize models trained on massive text corpora at relatively low monetary and computational costs. However, there are challenges associated with LLMs, particularly from biases observed in generated outputs \citep{feng_pretraining_2023, gupta_comprehensive_2024, motoki_more_2024, pit_whose_2024, ranjan_comprehensive_2024}.

A largely overlooked potential of LLMs lies in their capacity to act as comprehensive research assistants \citep{lu_ai_2024}. When equipped with predefined functions and tools, LLMs can perform a variety of repetitive, labor-intensive research tasks that extend beyond content generation or text analysis. A specific example of the research assistance capabilities of LLMs is information retrieval. LLMs can refine human-written queries, search through vast amounts of text, and retrieve contextually relevant results \citep{petroni_language_2019, hu_survey_2024, zhu_large_2023, dai_bias_2024, labruna_when_2024, zhang_agentic_2024}. However, traditional LLMs face limitations when it comes to handling field-specific contexts or addressing nuanced research questions. These issues often stem from the reliance of models like GPT \citep{radford_language_2019} and BERT \citep{devlin_bert_2018} on static, pre-trained knowledge bases. Without the ability to update their internal knowledge effectively, these models are prone to hallucinations, producing plausible but inaccurate or outdated information \citep{burger_use_2023, ji_survey_2023}. For example, \citet{finn_assessing_2024} found that ChatGPT struggled to provide accurate and up-to-date descriptions of political conditions across U.S. states, often relying on vague or misleading references.

\begin{figure}
    \centering
    \includegraphics[width=0.8\linewidth]{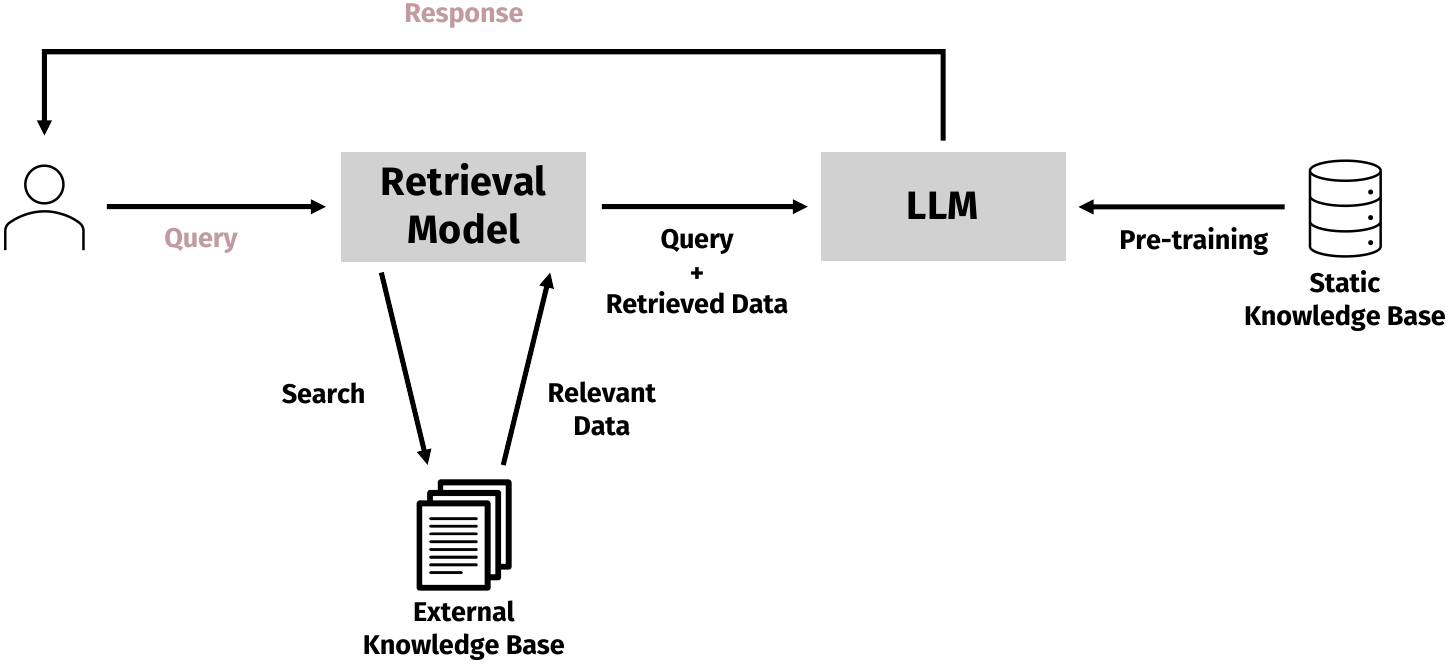}
    \caption{\textbf{Retrieval-Augmented Generation (RAG)}. RAG supplements a user's query with external information to allow an LLM to create output with up-to-date and contextually relevant content.}
    \label{fig:rag}
\end{figure}

Two approaches currently exist to address these limitations: fine-tuning \citep{ouyang_training_2022, schick_toolformer_2023} and retrieval-augmented generation (RAG) \citep{lewis_retrieval-augmented_2020}. Fine-tuning involves retraining a model on domain-specific data to update its internal knowledge, but this process is resource-intensive and risks overwriting existing capabilities. On the other hand, RAG, as depicted in \autoref{fig:rag}, dynamically retrieves external information in response to a user query, allowing the model to access up-to-date and contextually relevant content without model retraining. By incorporating external resources such as SQL databases, vector embeddings, or APIs, RAG enhances the precision, reliability, and transparency of LLM outputs \citep{thakur_beir_2021}.

\begin{figure}
    \centering
    \includegraphics[width=0.8\linewidth]{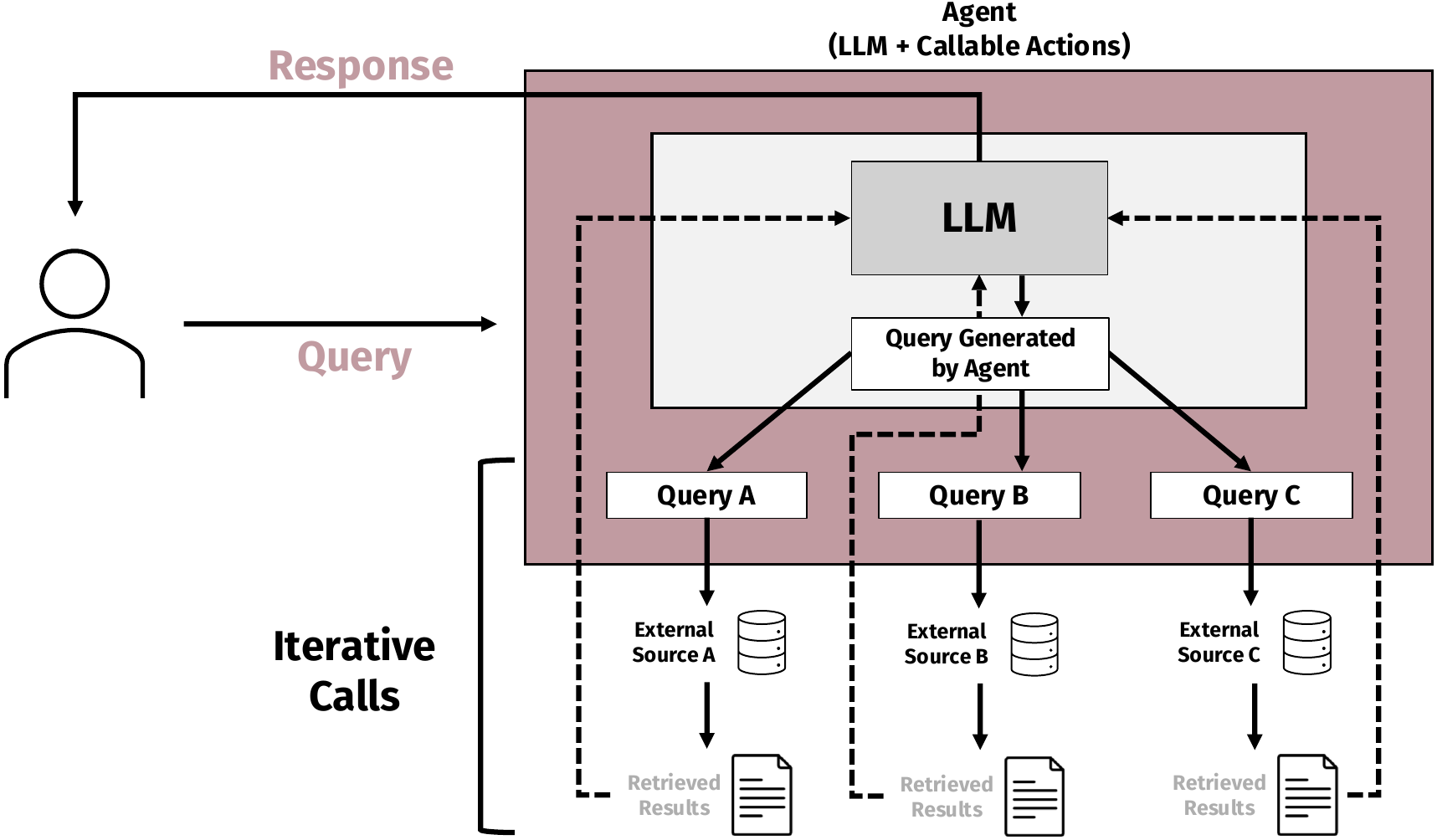}
    \caption{\textbf{Agentic RAG}. With a set of functions defined by the researcher and implementing an AI-assistant framework, LLMs can act as autonomous agents that dynamically decide when, where, and how to retrieve external information.}
    \label{fig:agentic-rag}
\end{figure}

As shown in \autoref{fig:agentic-rag}, while traditional RAG is an important enhancement, Agentic RAG builds on this capability by enabling LLMs to act as autonomous agents that dynamically decide when, where, and how to retrieve external information. Unlike traditional implementations of RAG that rely on predefined processes for data retrieval, LLMs equipped with an agentic interface can independently select the most appropriate data sources based on the task at hand. For example, within a single workflow, an LLM agent might query a vector database for semantic similarity, switch to a SQL database for structured information, or query APIs for real-time data. Moreover, an agent can iteratively refine its retrieval process, repeatedly querying different databases or the same source multiple times until it identifies plausible and persuasive information to address the user's query.

The flexibility of Agentic RAG provides an LLM with discretion in determining which resources to query and how to integrate retrieved data into the response given to the user. This dynamic approach ensures that the LLM agent can adapt its strategy based on context, using a combination of data sources to provide comprehensive and contextually accurate answers. This capability is particularly powerful in scenarios where the information required to respond to a user's request is fragmented across multiple sources or when a single data source alone cannot provide sufficient depth or breadth of knowledge. While Agentic RAG is a key feature of LLM agents, it represents just one of many functionalities that can be integrated into LLM agents. By combining dynamic retrieval with processing and analytical functions, the modular design of LLM agents has the capacity to streamline the research process across research contexts when designed to leverage shared tools and provided with task-specific instructions on how to use them. Whether retrieving, analyzing, or synthesizing data, these agents can significantly reduce manual effort required for scholarly work while maintaining consistency, reliability, and transparency in their outputs.

\section*{The Case for LLM Agents and the Study of Political Institutions}
Theoretical models are foundational to the study of political institutions. These models aim to be a reflection of the way the world operates, with a focus on how individuals, context, and organizational structures produce outcomes. As \citet{clarke_model_2012} note, models are tools that scholars make useful for framing lines of inquiry, grouping empirical patterns under one framework, speculating causal mechanisms, or deriving predictions. In constructing theoretical models, scholars strive to showcase their parsimony and boundedness, yet broad applicability. It is through empirical research designs that scholars probe implications of theoretical models on particular actors, organizational structures, and behavioral constraints. As \citet{ashworth_theory_2021} explain, with the right empirical design, scholars can adjudicate whether theoretical models are similar to what they aim to reflect or whether the implications of those models should be elaborated on, reinterpreted, or disentangled to better describe relevant mechanisms. To assess a theoretical model, researchers want to gather data on as many of its observable implications as possible. This involves collecting data across a variety of contexts. The more instances in which a model's implications are consistent with empirical realities, the more useful that model is \citep{king_designing_1994}.

We argue that LLM agents might assist researchers in overcoming the challenges associated with these endeavors. LLM agents can go beyond being mere search engines, acting as comprehensive research assistants \citep{lu_ai_2024}, capable of automating and streamlining tasks that are otherwise time-intensive and complex. Namely, scholars can leverage LLM agents to identify relevant case examples, locate or preprocess data sources, and construct measures. The study of political institutions often involves working with context-specific and evolving data, making traditional methods for data collection arduous and resource-intensive. For instance, identifying relevant historical case studies or compiling datasets often requires specialized knowledge and manual effort, such as reading and coding archival materials or synthesizing disparate data sources. LLM agents alleviate these burdens by integrating information retrieval, preprocessing, and synthesis into a cohesive, iterative workflow. They can dynamically select the appropriate tools and data sources, refine their outputs, and adapt their methods to the specific demands of the task at hand.

This adaptability makes LLM agents particularly valuable for political science research, where the upstream tasks of data collection and preparation often require repeated effort for different empirical designs or research contexts. By combining data retrieval capabilities with functions for data processing and analysis, LLM agents enable researchers to revisit and extend previous work with greater scalability. In the next section, we outline the important considerations researchers must take in designing and implementing LLM agents for their own purposes. To motivate the choices and tradeoffs researchers face, we describe how LLM agents can be useful for conducting research on the U.S. Congress, introducing our own LLM agent---which we call \texttt{CongressRA}---as a proof of concept. To demonstrate how LLM agents can assist researchers, we use \texttt{CongressRA} to revisit the work of \citet{binder_dynamics_1999}, which examines the dynamics of legislative gridlock. In doing so, we highlight how enabling an LLM to interact with external data sources allows scholars to efficiently complete multiple steps of the research process.

\section*{Designing LLM Agents}
An LLM agent is a system that combines a Large Language Model (LLM) with predefined tools. Within this framework, tools are implemented as callable functions written in code with clearly specified inputs and outputs which an LLM can interpret and autonomously decide to utilize  based on its understanding of the content and context of a user's request. To enhance controllability and effectiveness of LLM agents for specific use cases, fine-tuning is often required. By providing explicit system prompts or guidelines that instruct the agent on how to call functions in various scenarios, a researcher helps an LLM agent better understand the context of a problem and identify the most relevant functions. This process--known as short-shot learning or in-context learning \citep{brown_language_2020}--improves the agent's performance by expanding its capabilities with successful examples or strategies.

Implementing such agents can be achieved through frameworks like OpenAI's Assistant API or open-source solutions such as LangChain. These frameworks simplify the development of LLM agents by providing built-in mechanisms for defining functions, executing them, transmitting output, and handling the re-execution of functions when necessary. We recommend that those in political science wishing to develop an LLM agent for their purposes rely on these frameworks. Instead, the focus of researchers should be on key decisions such as selecting the most suitable LLM and constructing the specific functions their LLM agent can call on. Ultimately, it is the design of these functions that determines how effectively LLM agents can assist in the research process. 

Two principles provide a structured approach to designing and selecting functions. First, \textit{modularity}: functions should be designed as standalone components for seamless integration, removal, or updating as research needs evolve over time. That being said, functions should be constructed in a manner that recognizes the potential for how multiple functions can be used together to accomplish a task.  Second, \textit{reusability}: researchers should prioritize constructing functions that are broadly applicable across multiple projects within a specific area of research (e.g., functions for semantic search, database querying, and generating descriptive statistics).

A cornerstone capability for LLM agents is Retrieval-Augmented Generation (RAG), enabling them to access external knowledge bases and augment their responses with up-to-date and relevant information. Implementing RAG through predefined functions accessible to LLM agents involves careful planning of both data storage and the method used to query data. Determining which mode of data storage and querying is best suited for a researcher's own LLM agent depends on the types of data used in their work and the tasks they are looking to complete. For research in political science, there are three key approaches to implementing RAG functionality:
\begin{enumerate}
    \item \textbf{Structured Data Retrieval with SQL Databases}: For tasks requiring exact, structured outputs, relational databases are more appropriate. Important data sources from a researcher's field of study that are often available in tabular form should be stored in SQL databases, for example, to allow for easy retrieval. An LLM agent then can be designed to dynamically generate SQL queries necessary to retrieve information relevant to a user's request.
    \item \textbf{Semantic Search with Vector Databases}: Text-based data, such as news articles or legislative summaries, which cannot be effectively retrieved through traditional filtering or keyword-based querying, can be stored as vector embeddings (e.g., OpenAI embeddings). An LLM agent can be equipped with the capacity to perform semantic search---a way of searching through text that understands the meaning and phrases included in a search query---to retrieve entries that are contextually relevant to a user's request, even when exact matches are not present. For example, summaries of news articles published in the \textit{New York Times}, along with relevant metadata, could be stored in a vector database. Researchers could then search for articles related to specific policy issues based on related words and phrases rather than implementing a series of keyword searches.
    \item \textbf{Entity Relationships with Graph Databases}: Data depicting relationships between entities (e.g., connections between bills, lobbying groups, sponsors, and committee memberships) can be stored in graph databases in the form of nodes, edges, and properties. To interact with a graph database, an LLM agents can be equipped to perform Cypher queries in Neo4j.
\end{enumerate}

In explaining the ways RAG functionality can be included in the design of an LLM agent, we primarily focus on how to best retrieve information. The predefined actions of an LLM agent, however, can be constructed to include processes that help maintain and update the data sources the agent works with and to standardize information across data sources. Thus, an LLM agent can not only assists with data retrieval, but data management and integration as well. This ensures that a user's request is fulfilled with the most up-to-date and contextually relevant information from disparate data sources. 

Before demonstrating the utility of an LLM agent to the study of political institutions, it is important to underscore one key point: because of their reasoning capacity, LLM agents can dynamically chain function calls together and accomplish multi-step tasks \textit{without} user intervention. That means if a user's prompt would require both data retrieval and data analysis, an LLM agent can autonomously call on and retrieve the necessary output using relevant predefined functions. 

While the capabilities of LLM agents are extensive, the reliability with which LLM agents can perform these processes depends on thoughtful prompt design and usage by researchers. To that end, researchers should limit query complexity. AI assistant frameworks often have constraints on the number of function calls they can make within a single query. For example, OpenAI's function call limits typically allow up to five iterations per query. Complex prompts that require extensive iterations of function to complete may exceed these limits. Instead of overloading a single query, researchers should design workflows that split tasks across smaller, manageable prompts that are submitted to the LLM agent. This can be done by structuring agents to handle one row or unit of a task at a time. Then, a researcher can make multiple calls to the LLM agent programmatically using external tools like R or Python, varying the prompts so that each request targets a specific subtask. This approach allows the agent to work more robustly without becoming overloaded, ensuring reliability and scalability.

\section*{Example LLM Agent: \texttt{CongressRA}}
\begin{figure}
    \centering
    \includegraphics[width=.85\linewidth]{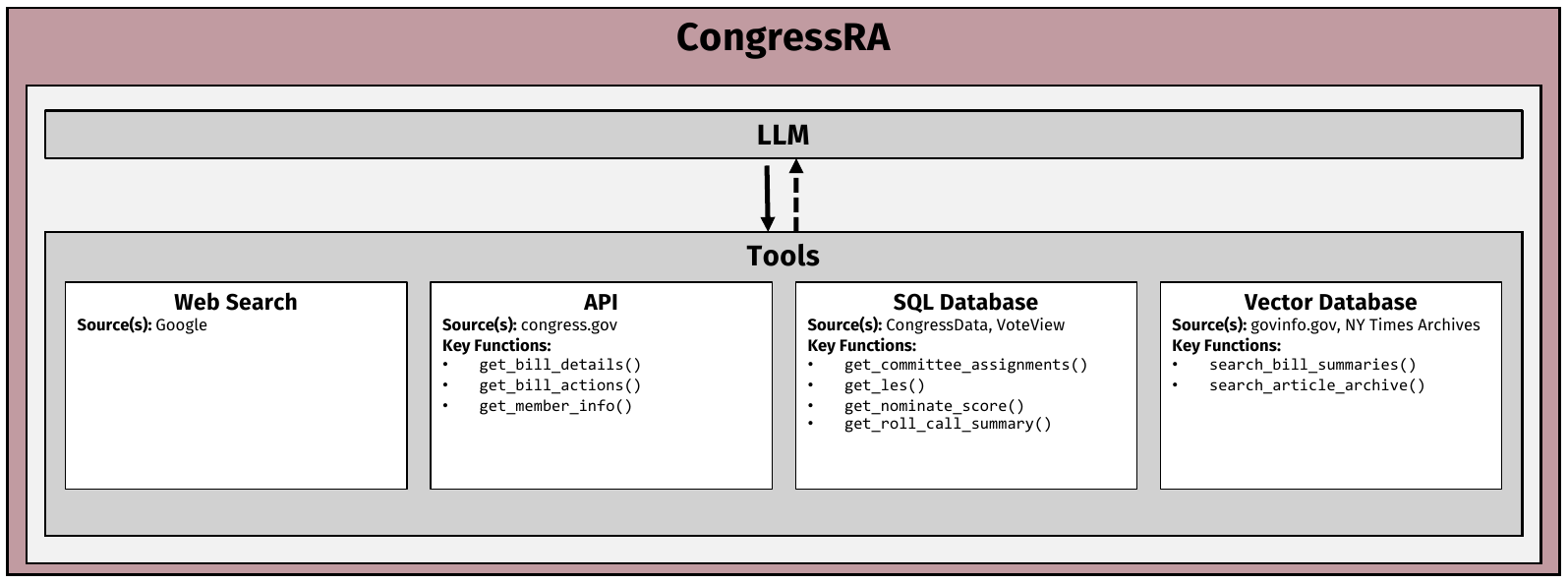}
    \caption{\textbf{CongressRA, an LLM agent for the study of the U.S. Congress}. There are four main sets of tools---along with the related external data sources---available to our LLM agent: web search, API access, SQL database querying, and vector database querying.}
    \label{fig:congressra}
\end{figure}

To showcase the potential of LLM agents in advancing research on political institutions, we present the design of \texttt{CongressRA}---a specialized agent for studying the U.S. Congress---as a proof of concept. Based on the design principles discussed earlier, \texttt{CongressRA} integrates an array of functions that interact with external data sources, retrieve structured and unstructured information, and process data. Two questions guided our design. First, which data sources do scholars of Congress rely on across most of the questions they aim to answer with empirical research designs? Second, from these data sources, what information do scholars typically extract? Answers to these questions guided how we made data sources accessible to our LLM agent and the input and output of each predefined action we set up for our LLM agent. The basic architecture of \texttt{CongressRA} is outlined in \autoref{fig:congressra}. A complete description of the data sources \texttt{CongressRA} interacts with and its predefined functions are described in Appendix \ref{sec:congresra_design}. 

The data sources \texttt{CongressRA} interacts with are regularly used by congressional scholars and provide the latest available information on members and legislative action within the U.S. Congress. To ensure our LLM agent can retrieve information about individual bills and members, we implemented functions to interact with the \texttt{congress.gov} API.\footnote{Specifically, we define functions to call the \texttt{GET /bill/:congress/:billType/:billNumber}, \texttt{GET /bill/:congress/:billType/:billNumber/actions}, and \texttt{GET /member/:bioguideId} endpoints. These enable the LLM agent to retrieve details related to individual bills typically required by researchers, a bill's legislative history, and basic demographic information about individuals.} We chose to have our LLM agent acquire data on bills and members via API calls because reliance on static CSV files opens up the possibility of retrieving outdated information. We supplement the data we can collect via the \texttt{congress.gov} API with tables from three data sources: \texttt{govinfo.gov}, \texttt{CongressData} \citep{grossmann_congressdata_2022} and \texttt{Voteview} \citep{lewis_voteview_2024}. We store these data in a remotely accessible SQL database to ensure they can be queried as needed through predefined functions. These predefined functions include the ability to retrieve information on the status of legislation from previous sessions of Congress, member committee assignments, member legislative effectiveness scores, member DW-NOMINATE scores, and records of individual roll call votes.\footnote{These data all retrieved from \citet{grossmann_congressdata_2022}. Committee assignment data are originally collected by \citet{stewart_iii_congressional_2017}. Legislative effectiveness scores (LES) were originally calculated by \citet{volden_legislative_2014}. DW-NOMINATE scores and roll call vote records were originally made available by \citet{lewis_voteview_2024}.} Lastly, we store text summaries of bills introduced in Congress (for over 80,000 bills introduced since 2012) and published articles (for nearly 700,000 articles published since 2012) from the \textit{New York Times} Archive API in a vector database and allow our LLM agent to perform semantic searches with these data.\footnote{Bill summaries were originally produced by \texttt{govinfo.gov}. The choice to incorporate information on new coverage based on articles from the \textit{New York Times} Archive API was arbitrary. In practice, a researcher can incorporate media data from any publication whose data is publicly accessible.} In doing so, we can use human-language queries to retrieve bills and articles rather than relying on keyword searches, thereby increasing the ability to return contextually-relevant results. 

The design choices we made in building \texttt{CongressRA} are consistent with the broader guidelines we have outlined in this paper for the successful use of Agentic RAG. In particular, our LLM agent interacts with data sources that are relevant for all those who study Congress and relies on predefined functions intended for specific subtasks and can be iterated on using external tools. By integrating semantic search (via vector databases) and precise querying (via SQL databases), \texttt{CongressRA} ensures both breadth and depth in its data retrieval capabilities. By leveraging multiple data sources, \texttt{CongressRA} effectively transforms labor-intensive research workflows into cohesive unit tasks. These tasks include, (1) tracing the legislative history of individual bills including their sponsors, any roll call votes taken, and how members of different types engaged in the legislative process; (2) assessing the level of media coverage associated with particular bills or policy debates; or (3) building a full dataset of member-level covariates from demographics to legislative participation. 

\subsection*{Application: Measuring Legislative Gridlock}
To demonstrate how LLM agents can streamline research, we replicate and the process \citet{binder_dynamics_1999} uses to measure legislative gridlock. Binder's measure of legislative gridlock responds to \citet{mayhew_divided_2005}, whose analysis of legislative productivity argues that, due to the incentives of individual members of Congress and prevailing political conditions, the enactment of significant legislation is always possible, regardless of whether there is unified or divided government. Binder reevaluates this theoretical argument by suggesting that to measure how capable Congress is in enacting policies that addresses pressing issues, all possible policies that \textit{could} have been enacted---not just those that were enacted---need to be factored into a measure of legislative productivity. Thus, Binder's measure of legislative gridlock is the proportion of all issues perceived as requiring public attention (i.e., the "systemic agenda") that fails to see relevant policy enacted into law.

Constructing Binder's measure of legislative gridlock involves several labor-intensive steps: (1) identify all salient policy issues within a session; (2) identify proposed bills related to each policy issue in a given session; and (3) determine whether these bills were enacted into law. Step (1) required Binder and their team of researchers to read over 15,000 unsigned editorials published in the \textit{New York Times} between 1947 and 1996, cataloging policy issues mentioned in each of them. For Step (2), Binder identified proposed legislation related to these policy using a variety of sources, including the \textit{Congressional Quarterly Almanac} and the Library of Congress's website. In step (3), Binder uses these sources to determine whether any legislation related to each policy issue was enacted into law.

\begin{figure}
    \centering
    \includegraphics[width=\linewidth]{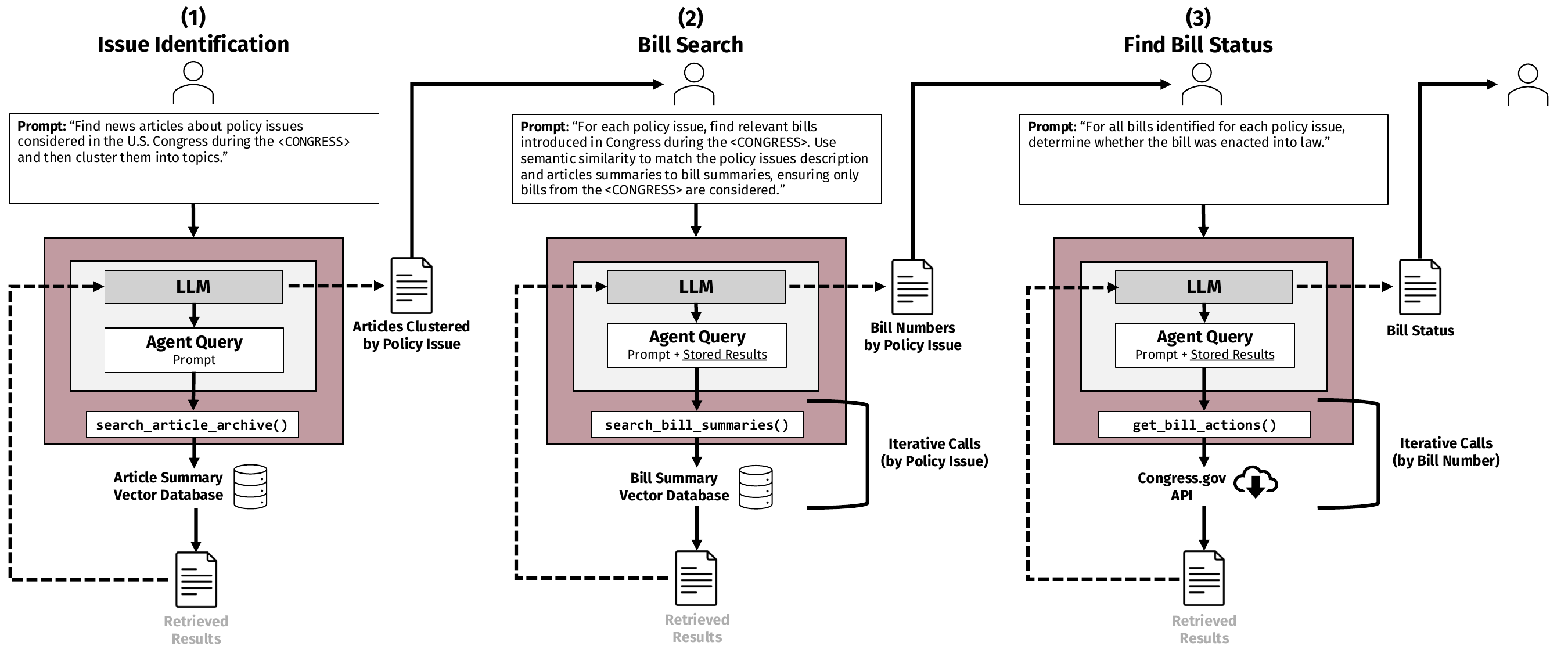}
    \caption{\textbf{Using \texttt{CongressRA} to measure legislative gridlock}. To follow the methodology \citet{binder_dynamics_1999} follows, a user submits three prompts to the LLM agent to identify salient policy issues, search for relevant legislation, and identify if any such legislation was enacted into law. This process is repeated to produce a value of legislative gridlock in each session of Congress.}
    \label{fig:binder}
\end{figure}

The process we follow using \texttt{CongressRA} to mimic Binder's methodology is illustrated in \autoref{fig:binder}. To evaluate how levels of legislative gridlock have changed in recent years, we repeat this process for all two-year periods from 2013 to 2024, covering the 113th through 118th Congress. In step (1), we submit a prompt to the LLM agent asking it to identify salient policy based on news articles published in the desired year. In building \texttt{CongressRA}, we used the \textit{New York Times} Article Search API to collect summaries of all articles published since 2012, converted their headlines and summaries into vector embeddings via the OpenAI embedding API, and stored these vector embeddings in a vector database. With the \texttt{search\_article\_archive()} function, \texttt{CongressRA} is then capable of searching across the more than 100,000 articles available in each congress using human-language query rather than a keyword search.\footnote{To be clear, this is a deviation from the process Binder followed. Rather than searching through only unsigned editorials, our process searches through all published articles appearing in the \textit{New York Times} in a given year. This deviation, however, was a limitation of how extractions of article information via the \textit{New York Times} Article Search API can be made.} The submitted prompt requires the agent to perform multiple actions sequentially: retrieve relevant articles and cluster or classify them by topic. To complete this process, the agent begins by transforming the query into a vector embedding to search the vector database for relevant articles. The agent then reads the entries returned from the vector database and organizes them into clusters based on common themes and topics within their content. Notably, this second action does not involve unsupervised machine learning algorithms--the process of organizing unstructured data into categories is one that the LLM incorporated into our agent is capable of performing on its own. 

In response to the prompt from step (1), the LLM agent provides a summary and list of associated articles for each policy issue it discovers. An example of what the output looks like for a single policy issue is shown in Appendix \ref{sec:cluster_example}. In the 116th Congress (2019-2020), our LLM agent identifies 10 distinct policy issues: "Government Funding and Shutdowns", "Climate Change", "Gun Control", "COVID-19 Response and Relief", "Police Reform and Racial Justice", "War Powers and Foreign Policy", "Tech Regulation and Privacy", "Congressional Procedure and Reform", "Healthcare Policy", and "Presidential Impeachment." Overall, step (1) illustrates a common flow in Agentic RAG where an agent calls functions, obtains results from those functions, and generates a response to the user's prompt based on the obtained results.

In step (2), we submit a prompt to the LLM agent asking it find relevant bills introduced in Congress during the specified session that correspond to each of the policy issues identified in step (1). Similar to our choice on how to store and search news article information, summaries of legislation produced by \texttt{govinfo.gov} were converted into vector embeddings embeddings and stored in a vector database. In total, the vector database contains summaries of over 80,000 bills introduced since 2012. This allows the LLM agent to perform semantic similarity searches---as part of the \texttt{search\_bill\_summaries()} function---using its own query written based on its understanding of the prompt we submitted and the results from step (1) stored in memory. In response to the prompt, the agent returned a list of bills for each policy issue, in descending order of cosine similarity scores.\footnote{The ranking of semantic search results retrieved from a vector database is a typical feature of vector databases. More information on how semantic search results are presented can be found at \href{https://www.pinecone.io/learn/what-is-similarity-search/}{\texttt{https://www.pinecone.io/learn/what-is-similarity-search/}}.} An example of this output is found in Appendix \ref{sec:relevant_leg}.

Lastly, in step (3), we submit a prompt to the LLM agent asking it to determine if any legislation identified in step (2) was enacted into law. To do so, \texttt{CongressRA} autonomously calls the \texttt{get\_bill\_status()} function to query a table in a SQL database containing an indicator of whether each piece of legislation was enacted into law. The agent used the bill ID (comprising the session of Congress, bill type, and bill number) to formulate the necessary query. The agent automatically makes iterative calls to this function to check the enactment status of all legislation identified in step (2). In the 116th Congress, based on the agent's findings and our own manual review, among the 10 policy issues identified in step (1), 6 policy issues had at least one corresponding piece of legislation enacted into law. These issues included: "Government Funding and Shutdowns," "COVID-19 Response and Relief," "War Powers and Foreign Policy," "Tech Regulation and Privacy," "Congressional Procedure and Reform," and "Healthcare Policy." 

\begin{figure}
    \centering
    \includegraphics[width=0.6\linewidth]{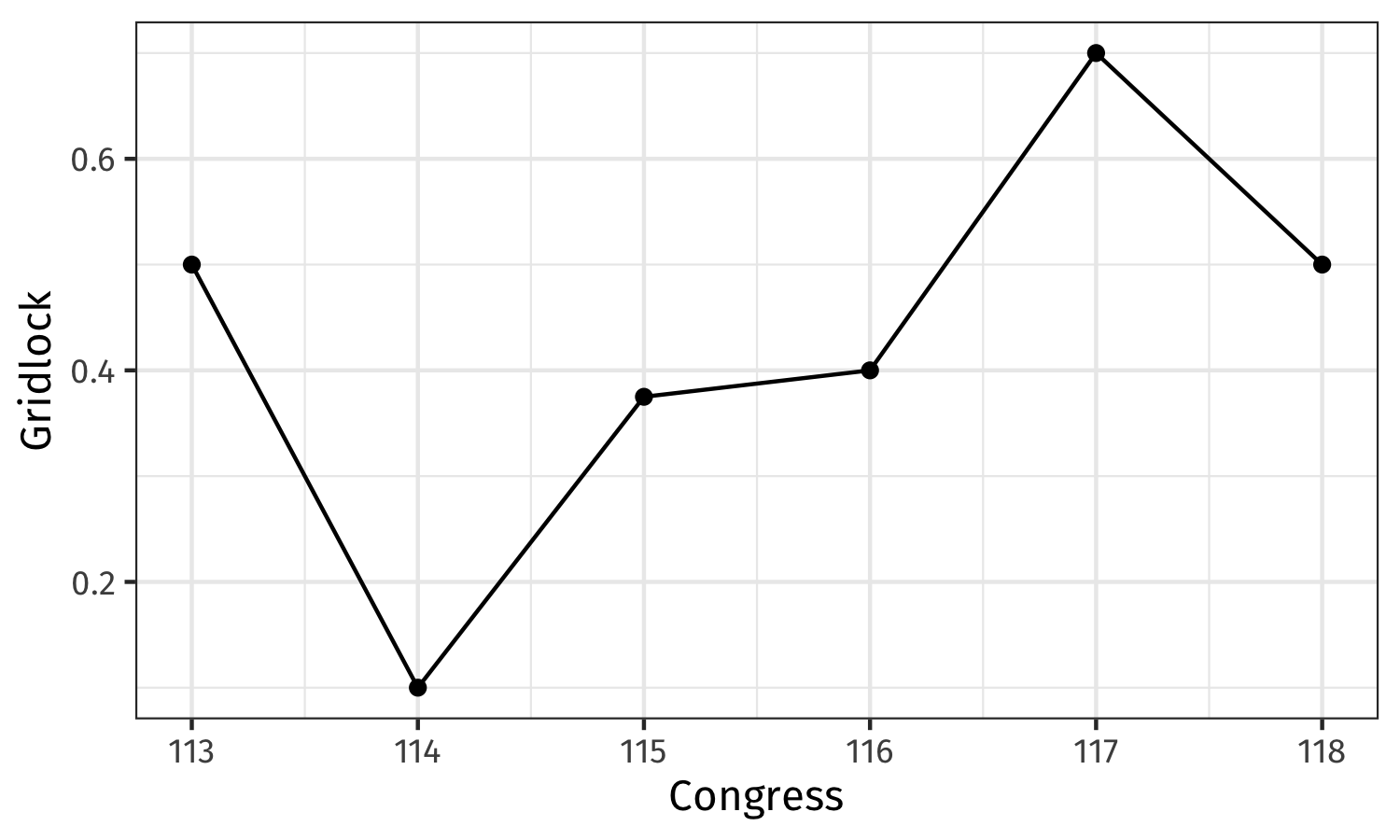}
    \caption{\textbf{Legislative Gridlock (113th-118th Congress)}}
    \label{fig:gridlock}
\end{figure}
Thus, in the 116th Congress, the legislative gridlock score---calculated as the proportion of salient policy issues that were not addressed through enacted legislation---is 0.40. This indicates that approximately 40\% of salient policy issues were not addressed through enacted legislation during the 116th Congress. In \autoref{fig:gridlock}, we present the values of legislative gridlocked produced by \texttt{CongressRA} for the 113th through 118th Congress. A complete benchmarking of the performance of \texttt{CongressRA} compared to using other off-the-shelf LLMs (e.g., \texttt{Claude-3-7-extended-reasoning model}, \texttt{Perplexity.ai::Deep-Research} and \texttt{OpenAI::Deep-Research}) for following Binder's methodology to measure legislative gridlock is found Appendix \ref{sec:benchmark}. While \texttt{CongressRA} identifies a similar set of policy issues important to the 113th Congress and shows similar patterns of legislative gridlock in the 113th through 118th Congress compared to other models, only the LLM agent provides the transparency and detail needed for researchers to trust these tools. \texttt{CongressRA} offers a complete list of news articles used to identify policy issues and a full catalog of related bills. Other models, due to token limits and comparatively less comprehensive information retrieval capabilities, do not provide this level of detail for users to verify the accuracy of gridlock values.

\subsection*{Reflecting on the Performance of \texttt{CongressRA}}
While our approach is not identical to Binder's original method, it follows the same spirit by identifying salient policy issues and assessing legislative action on them. However, there are some notable differences. First, whereas we searched through all articles published in the \textit{New York Times}, Binder and their research team manually searched only through unsigned editorials to identify salient policy issues. As a result, our approach likely identifies are broader set of policy issues. This is a key advantage of our approach though: what was previously a time- and resource-intensive tasks is now automated, completing the process of measuring legislative gridlock across sessions with minimal additional effort. Second, because of data availability, the values of legislative gridlock that we produce cover a different time period than the period of 1947 to 1996 that appears in Binder's original article. To be clear, the values of legislative gridlock we provide in this paper are intended to highlight the scalability and efficiency LLM agents like \texttt{CongressRA} offer rather than be a replication and reanalysis of Binder's findings.

Even when using an LLM agent like \texttt{CongressRA} for data collection, processing, and measurement, there will still be points in which researchers need to make substantive decisions. For example, when performing semantic searches using vector embeddings, it is crucial to set appropriate thresholds for similarity scores to ensure only relevant results are returned. Cosine similarity scores range from $-1$ to $1$, with higher scores indicating greater similarity. In our bill search, the agent retrieved bills in descending order of similarity scores. However, we found that the agent did not always set satisfactory thresholds to filter out unrelated bills. To maintain the accuracy of the measure, we still needed to manually review lists of retrieved bills for each policy issue and set appropriate thresholds to include only those bills directly related to the policy area.

Furthermore, because commercial APIs used to implement LLMs often have token limits for inputs and outputs, developers must set up the predefined actions their LLM agent can take in a way that accounts for these limits.\footnote{A token is a unit of text used by language models for processing and generation. Tokens can be as short as a single character or as long as an entire word or phrase, depending on the tokenization scheme. Token limits refer to the maximum number of tokens that can be processed in a single input or output by the model.} In structuring \texttt{CongressRA}'s vector search capability for news articles, we accommodated token limits by reducing the maximum number of articles retrieved in a single vector search (i.e., lowering the \textit{top K} value) to 200 articles.

Lastly, agents should be equipped with error-handling logic to execute functions effectively. In designing \texttt{CongressRA}, we ensured the agent could identify which part of the process caused an error by analyzing error logs and either automatically adjust its parameters or explain the issue. This self-correcting mechanism is a notable advantage of the recent surge in LLM agents across various domains, making them more flexible and less error-prone.

\section*{Further Potential Applications}
By showing how \texttt{CongressRA} assists in measuring legislative gridlock across years, we presented just one potential application of LLM agents in the study of political institutions. There are numerous other examples, one of which is the study of policy diffusion. In the American context, a key mechanism for the distinct patterns in policy diffusion across states is increased coordination among activists, interest groups, and policymakers to create cohesive state-level policy agendas consistent with each party's brand \citep{grossback_ideology_2004, dellavigna_policy_2022, grumbach_laboratories_2022}. Actions of groups like the American Legislative Exchange Council (ALEC)---supporting legislation on issues ranging from economic development to voter ID requirements---show that organizations are particularly crucial to understanding the diffusion and polarization of state policy \citep{hertel-fernandez_state_2019}. Producing quantitative measures of policy diffusion and linkage---especially ones relying on legislative language---are difficult to produce. An existing approach is to identify "model" legislation created by organizations, collect a large corpora of legislative text, and implement algorithms for detecting text reuse \citep[e.g, ][]{burgess_legislative_2016,linder_text_2020}. By storing legislative text in a vector database, equipping an LLM agent with semantic search capability to identify legislation of interest, and creating functions for the agent to call that produce values of text similarity estimators, a researcher can formulate a series of prompts to produce evidence of policy diffusion more efficiently than ever before.

More applicable to those who study the U.S. Congress might be leveraging the capabilities of LLMs to transcribe video into text as one modular component of a larger LLM agent like \texttt{CongressRA}. Through this, a user could transcribe specific legislative committee hearings and create measurements about their content like the process \citet{ban_bureaucrats_2025} follow. Another similar use-case would be including predefined functions for retrieving videos of floor speeches, transcribing them, and then adding them to a vector database of floor speeches. Given the existing capabilities of \texttt{CongressRA}, this would be make it simple for a researcher to identify a set of bills, identify which members introduced them, and search for how members have justified similar proposals on the legislative floor in the past. 

The key point is that any of these extensions involve adding a few additional functions to an already well-designed system constructed to help those sharing similar research agendas. These LLM agents already have access to relevant databases or APIs, functions for querying data from these sources, and an integration with an LLM. To tailor the system for more pointed research questions involves smaller tweaks, like adding functions for implementing specific measures, processing data in a desired manner, or retrieving output of a specified structure.

\section*{Discussion}
As scholars of political institutions, the data we work with is specific to the context we study. As theories are developed and then tested with empirical research designs, we collect data on the individuals and the structures we study. While these theories aim to be generalizable, their utility often hinges on their ability to adapt to new contexts and remain robust when confronted with additional variables. It is through the iterative process of collecting and analyzing new data that we evaluate the transportability of theoretical models and refine their applicability to broader political phenomena.

An impediment, however, is that relevant data are frequently found and maintained across disparate sources, some of which are outdated or inconsistently structured, making integration labor-intensive and prone to error. While LLMs have recently entered the political science research toolkit primarily for content generation and text analysis, this paper has argued and demonstrated that their potential extends beyond these purposes. When enhanced with predefined functions and integrated with domain-specific data sources, LLM agents are uniquely positioned to assist researchers in the complex and time-intensive tasks of data collection, preprocessing, and analysis. The ability of \texttt{CongressRA}---our example LLM agent---to mirror the process of constructing \citet{binder_dynamics_1999}'s legislative gridlock index with a fraction of the effort traditionally required highlights the efficiency gains these tools can offer. By automating data collection processes and integrating diverse data sources, \texttt{CongressRA} demonstrated that LLM agents can extend the temporal and substantive scope of foundational studies in a scalable manner.

The utility of LLM agents, however, depends on thoughtful design and implementation. Developers must make deliberate choices about data storage and actions the agent will perform. Two critical considerations emerge for developers aiming to create tools that benefit a broad community of scholars.

First, data accessibility must be prioritized. Researchers rely on timely, structured, and contextually relevant data, and the way data is stored and retrieved significantly impacts an LLM agent's utility. For instance, \texttt{CongressRA} demonstrated how SQL databases can efficiently handle structured, tabular data, while vector databases excel in semantic search for unstructured text. The ability to integrate APIs and conduct web searches further ensures that LLM agents can access the most up-to-date and comprehensive information available.

Second, to best assist researchers in all phases of the empirical process---data collection, processing, and analysis---developers should be intentional in creating predefined functions that are parsimonious, targeted to specific subtasks, iterable, and most importantly, reusable across research contexts. For \texttt{CongressRA}, we implemented functions with broad use cases that retrieve specific important data points as well as specialized functions to process data and produce measurements. This modular approach not only streamlines the development process but also ensures that agents can evolve with changing research needs, maintaining their relevance over time.

We would be remiss in failing to acknowledge a fundamental question from those skeptical of using LLMs in tasks that are fundamental to the research process. LLMs have been critiqued for their lack of transparency. Largely developed by private firms, most researchers have little insight into the mechanics of how a user's prompt yields specifics outputs among the menu of LLMs researchers can choose from. While an LLM agent like \texttt{CongressRA} can be given clear instructions of when and how to utilize the predefined functions the architect of the LLM agent provides, the actual LLM that is the core of the agentic framework still maintains its own reasoning capacity in evaluating when and how to best use predefined functions. Recently, however, some available LLMs have made it possible for end users to see the thought process a model completes in transforming user prompts into outputs \citep[e.g., ][]{deepseek-ai_deepseek-r1_2025}. As advances continue to be made, researchers will hopefully come to expect and rely on this level of transparency to trust LLM agents as comprehensive research assistants.

To be clear, successfully creating an LLM agent that can assist scholars sharing common research interests is technically-demanding. The integration of LLM agents into political science research, however, represents an opportunity to address long-standing barriers to beginning research on particular institutions. As the field embraces these technologies, fostering a community-driven approach to the development and refinement of LLM agents will be critical in ensuring their impact.

	\newpage
	\singlespacing
	  \bibliographystyle{apalike}
    \bibliography{references}
    \clearpage
    \appendix 
    \renewcommand\thetable{\thesubsection-\arabic{table}} 
    \renewcommand\thefigure{\thesubsection-\arabic{figure}} 
    \renewcommand\thesubsection{\Alph{subsection}} 
    
    \renewcommand{\thepage}{A-\arabic{page}} 
    \setcounter{page}{1}
    \setcounter{figure}{0}
    \setcounter{table}{0}
    \setcounter{footnote}{0}
    \singlespacing

    \section*{Online Appendix for "Agent-Enhanced Large Language Models for Researching Political Institutions"}
    \vspace{5mm}
    \localtableofcontents
    \etocdepthtag.toc{mtappendix}
    \etocsettagdepth{mtchapter}{none}
    \etocsettagdepth{mtappendix}{subsection}
    \clearpage
    \subsection{Design of \texttt{CongressRA}}
    \label{sec:congresra_design}
    \subsubsection{Data Sources and Key Functions}
    The following outlines the primary data sources integrated into \texttt{CongressRA}, the functions implemented to interact with these sources, and specific research questions these functions are designed to address.
    
    \begin{itemize}
        \item \textbf{Data Source: \texttt{congress.gov API}}
        \begin{itemize}
            \item \textbf{Purpose:} Retrieve comprehensive information about Congressional bills, their legislative histories, sponsors, and related subjects. \texttt{CongressRA} is given the capabilities to use functions that directly implement specific \texttt{congress.gov} API endpoints.
            \item \textbf{Key Functions}:
            \begin{itemize}
                \item \texttt{get\_bill\_details}: Fetches essential details about a bill, such as its title, sponsors, and introduction date. This implements the \\\texttt{GET /bill/:congress/:billType/:billNumber} endpoint.
                \begin{itemize}
                \item \textbf{Input:} \texttt{bill\_id}, a unique identifier for each bill, consisting of Congress number (e.g., 118), bill type (e.g., H.R.), and bill number (e.g., 5376).
                \item \textbf{Output:} Bill title, sponsor information (e.g., bioguide IDs), introduction date.
                \item \textbf{Example Query:} \textit{"What are the key details of the Postal Service Reform Act of 2022?"} 
                \end{itemize}
            
            \item \texttt{get\_bill\_actions}: Retrieves the legislative history of a bill, including major actions taken and their dates. This implements the \\\texttt{GET /bill/:congress/:billType/:billNumber/actions} endpoint.
            \begin{itemize}
                \item \textbf{Input:} \texttt{bill\_id}.
                \item \textbf{Output:} List of legislative actions (e.g., "Passed House on 2022-02-08").
                \item \textbf{Example Query:} \textit{"What steps were taken to pass the Affordable Care Act in Congress?"}
            \end{itemize}
            \item \texttt{get\_member\_info}: Provides biographical and Congressional role information for individual members of Congress. This implements the \\\texttt{GET /member/:bioguideid} endpoint.
            \begin{itemize}
                \item \textbf{Input:} \texttt{bioguide\_id}, a unique identifier for each member of Congress.
                \item \textbf{Output:} Member name, state, chamber, leadership positions, terms served.
                \item \textbf{Example Query:} \textit{"What leadership roles has Senator Elizabeth Warren held in Congress?"}
            \end{itemize}
            \end{itemize}
        \end{itemize}
        \item \textbf{Data Source:} \texttt{CongressData} \citep{grossmann_congressdata_2022}
        \begin{itemize}
            \item \textbf{Purpose:} Retrieve supplementary member information, such as committee assignments and legislative effectiveness scores. Given that these data are available in table form, they are downloaded and made available to \texttt{CongressRA} in a SQL database.
            \item \textbf{Key Functions}:
            \begin{itemize}
                \item \texttt{get\_committee\_assignments}: Identifies the committees a member served on during a specific Congress and whether they held a chair position.
                \begin{itemize}
                    \item \textbf{Input:} \texttt{bioguide\_id}, Congress.
                    \item \textbf{Output:} Committees served on, indicators for chair roles.
                    \item \textbf{Example Query:} \textit{"Which committees did Senator John McCain chair during the 108th Congress?"}
                \end{itemize}
                \item \texttt{get\_les}: Retrieves the Legislative Effectiveness Score (LES) for a member, reflecting their legislative productivity.
                \begin{itemize}
                    \item \textbf{Input:} \texttt{bioguide\_id}, Congress.
                    \item \textbf{Output:} LES value.
                    \item \textbf{Example Query:} \textit{"How effective was Representative Alexandria Ocasio-Cortez during the 117th Congress?"}
            \end{itemize}
            \end{itemize}
            
        \end{itemize}
        \item \textbf{Data Source:} \texttt{Voteview} \citep{lewis_voteview_2024}
        \begin{itemize}
            \item \textbf{Purpose:} Retrieve NOMINATE scores for ideological analysis and detailed roll-call vote information. Given that these data are available in table form, they are downloaded and made available to \texttt{CongressRA} in a SQL database.
            \item \textbf{Impact:} Researchers analyzing voting patterns and ideological trends can automate data retrieval and focus on analysis, avoiding time-consuming manual data extraction.
            \item \textbf{Key Functions:}
            \begin{itemize}
            \item \texttt{get\_nominate\_score}: Fetches NOMINATE scores for ideological placement on key dimensions.
            \begin{itemize}
                \item \textbf{Input:} \texttt{bioguide\_id}, Congress.
                \item \textbf{Output:} First and second dimension NOMINATE scores.
                \item \textbf{Example Query:} \textit{"How has Senator Mitt Romney’s ideological position shifted over his Congressional career?"}
            \end{itemize}
            \item \texttt{get\_roll\_call\_summary}: Provides summaries of roll-call votes, including yea/nay counts and vote outcomes.
            \begin{itemize}
                \item \textbf{Input:} \texttt{bill\_id}.
                \item \textbf{Output:} Roll number, yea/nay counts, vote result.
                \item \textbf{Example Query:} \textit{"What were the vote totals for the Infrastructure Investment and Jobs Act?"}
            \end{itemize}
        \end{itemize}
        \end{itemize}
        \item \textbf{Data Source:} \textit{New York Times} API
        \begin{itemize}
            \item \textbf{Purpose}: Provide \texttt{CongressRA} with a searchable database of published articles. For enabling semantic search, article summaries and metadata are stored in a vector database. The database is populated through a series of calls to the Archive API endpoint covering the years 1997 through 2024.
            \item \textbf{Key Function}:
            \begin{itemize}
                \item \texttt{search\_article\_archives()}: Runs a semantic query to a vector database.
                \begin{itemize}
                    \item \textbf{Input}: human-language search query
                    \item \textbf{Output}: JSON of article metadata and summary
                    \item \textbf{Example Query}: \textit{Retrieve \textit{New York Times} articles that discuss debates on abortion policy in the U.S. Congress in 2015}.
                \end{itemize}
            \end{itemize}
        \end{itemize}
        \item \textbf{Data Source}: \texttt{govinfo.gov}
        \begin{itemize}
            \item \textbf{Purpose}: Provide \texttt{CongressRA} with summaries and enactment status of introduced legislation. Enactment status is collected from the \href{https://www.govinfo.gov/bulkdata/BILLSTATUS}{Bill Status Bulk Data Repository}. Legislation summaries are collected from the \href{https://www.govinfo.gov/bulkdata/BILLSUM}{Bill Summary Bulk Data Repository}. Both data points are added to a vector database.
            \item \textbf{Key Functions}:
            \begin{itemize}
                \item \texttt{search\_bill\_summaries()}: Runs a semantic search query to a vector database. \begin{itemize}
                    \item \textbf{Input}: human-language search query
                    \item \textbf{Output}: JSON of bill data metadata and summary
                    \item \textbf{Example Query}: \textit{Retrieve bills introduced in the U.S. Congress in 2016 related to election administration}.
                \end{itemize}
            \end{itemize}
        \end{itemize}
    \end{itemize}

        \subsubsection{Methods of Data Storage and Query: Vector and SQL Databases}
        \begin{itemize}
            \item \textbf{Purpose:} Enhance Retrieval-Augmented Generation (RAG) with tailored querying capabilities.
            \item \textbf{Key Functionalities}:
            \begin{itemize}
            \item \textbf{Vector Database for Semantic Search:} Suitable for unstructured data. By embedding text into vectors, the agent can retrieve semantically similar content efficiently. As an example, we create a vector database with summaries of articles published in the \textit{New York Times} and summaries of legislative bills introduced in the U.S. Congress.
            \begin{itemize}
                \item \textbf{Example Query:} \textit{"Find editorials in the New York Times discussing Congressional health care reform from 2010. Then find bills introduced in Congress related to this news coverage"}
            \end{itemize}
            \item \textbf{SQL Database for Structured Queries:} We store data from the \texttt{CongressData} and \texttt{Voteview} in a SQL database. The functions mentioned above run the relevant queries to extract the desired data. This is ideal for retrieving exact information, such as committee memberships or roll-call vote records.
            \begin{itemize}
                \item \textbf{Example Query:} \textit{"Who were the members of the Senate Banking Committee during the 117th Congress?"}
            \end{itemize}
        \end{itemize}
        \end{itemize}

    \subsection{\texttt{CongressRA} Output}
    \subsubsection{Example of Policy Issue Identification}
    \label{sec:cluster_example}
    In the 113th Congress (2013-2014), \texttt{CongressRA} analyzes 200 articles from the \textit{New York Times} and provides the following outputs to explain the policy issues it identifies. The summary output displays the policy issue name, a summary of what news articles reveal about the policy issues, and the query the LLM generated that will be used to search for bills using the vector database of bill summaries.

    \begin{tcolorbox}[%
    enhanced, 
    breakable,
    skin first=enhanced,
    skin middle=enhanced,
    skin last=enhanced,
    ]
        \textbf{Summary Output}:
        \begin{enumerate}
        \item Government Funding and Fiscal Showdowns
        \begin{itemize}
            \item Articles: 18
            \item Summary: The 113th Congress was marked by repeated fiscal crises, including a 16-day government shutdown in October 2013 and multiple debt ceiling confrontations. Budget negotiations were characterized by brinkmanship and last-minute deals.
            \item Query: \texttt{government shutdown debt ceiling budget appropriations fiscal crisis}
        \end{itemize}

        \item Immigration Reform
        \begin{itemize}
            \item Articles: 12
            \item Summary: Despite early bipartisan momentum and Senate passage of comprehensive immigration reform, the legislation stalled in the House, eventually leading to executive action by President Obama in late 2014.
            \item Query: \texttt{immigration reform pathway citizenship border security executive action}
        \end{itemize}

        \item Gun Control Legislation
        \begin{itemize}
            \item Articles: 14
            \item Summary: Following the Newtown school shooting, Congress considered various gun control measures including expanded background checks, but despite initial momentum, key legislation failed in the Senate in April 2013.
            \item Query: \texttt{gun control background checks assault weapons Newtown filibuster}
        \end{itemize}

        \item Congressional Dysfunction and Procedure
        \begin{itemize}
            \item Articles: 22
            \item Summary: The 113th Congress was historically unproductive, passing fewer laws than almost any modern Congress. Articles highlight procedural changes like filibuster reform and the shift of policymaking to executive agencies and courts.
            \item Query: \texttt{congressional dysfunction filibuster reform productivity gridlock polarization}
        \end{itemize}

        \item Foreign Policy and National Security
        \begin{itemize}
            \item Articles: 11
            \item Summary: Congress grappled with its role in foreign policy decisions regarding Iran sanctions, Syria intervention, and NSA surveillance programs, often with partisan divisions complicating the administration's diplomatic efforts.
            \item Query: \texttt{foreign policy Iran sanctions Syria authorization NSA surveillance}
        \end{itemize}

        \item Healthcare and Affordable Care Act
        \begin{itemize}
            \item Articles: 9
            \item Summary: The Republican-controlled House repeatedly passed bills to repeal or defund the Affordable Care Act, while the Democratic Senate blocked these efforts. The government shutdown of 2013 was largely triggered by this standoff.
            \item Query: \texttt{Affordable Care Act Obamacare repeal defund healthcare reform}
        \end{itemize}

        \item  Executive Action and Congressional Authority
        \begin{itemize}
            \item Articles: 10
            \item Summary: As legislative gridlock persisted, President Obama increasingly turned to executive actions to advance his agenda, raising questions about the balance of power between branches and the limitations of such approaches.
            \item Query: \texttt{executive action executive order presidential power congressional authority}
        \end{itemize}

        \item Tax and Economic Policy
        \begin{itemize}
            \item Articles: 8
            \item Summary: Despite bipartisan interest in tax reform, Congress made little progress beyond temporary extensions of existing tax provisions. Economic policy initiatives like minimum wage increases and unemployment benefits faced partisan gridlock.
            \item Query: \texttt{tax reform economic policy minimum wage tax extenders}
        \end{itemize}
        
    \end{enumerate}
    \end{tcolorbox}

    \textbf{JSON Output}:
    \begin{lstlisting}
        "clusters": [
        {
          "name": "Government Funding and Fiscal Showdowns",
          "articles": [
            "The 113th Congress is concluding with partisan brinksmanship and one last mad scramble 
            for votes to pass a $1.1 trillion spending package.",
            "The near-death experience of a bipartisan, governmentwide funding bill on Thursday 
            highlighted a fundamental problem with Congress at the moment: the lost art of compromise.",
            "Hours after the Senate passed its measure, the House voted 285 to 144 to approve 
            the Senate plan, which would fund the government through Jan. 15 and raise the debt limit 
            through Feb. 7."
          ],
          "article_count": 18,
          "summary": "The 113th Congress was marked by repeated fiscal crises, including a 16-day 
          government shutdown in October 2013 and multiple debt ceiling 
          confrontations. Budget negotiations were characterized 
          by brinkmanship and last-minute deals.",
          "query": "government shutdown debt ceiling budget appropriations fiscal crisis"
        },
        {
          "name": "Immigration Reform",
          "articles": [
            "Speaking at an immigration event in Nashville, President Obama urged Congress 
            to pass comprehensive immigration legislation.",
            "A bipartisan group in the House of Representatives has been meeting sporadically to 
            create a comprehensive immigration bill, but the Senate debate 
            is dwarfing the group's efforts.",
            "With immigration legislation dead for the year, Congress has a very short must-do list 
            as relations between the two parties, already miserable, seem to be 
            getting worse."
          ],
          "article_count": 12,
          "summary": "Despite early bipartisan momentum and Senate passage of comprehensive 
          immigration reform, the legislation stalled in the House, eventually 
          leading to executive action by President Obama in late 2014.",
          "query": "immigration reform pathway citizenship border security executive action"
        },
        {
          "name": "Gun Control Legislation",
          "articles": [
            "Congress is beginning to look at options for gun safety legislation after 
            the Newtown shooting.",
            "The committee will begin to consider four separate gun safety measures on Thursday, 
            although the process of narrowing actual legislative language 
            will most likely dribble into next week.",
            "Talks to revive legislation are quietly under way as a bipartisan group seeks 
            a way to bridge the differences that led to last week's collapse 
            of a background check bill."
          ],
          "article_count": 14,
          "summary": "Following the Newtown school shooting, Congress considered various gun control 
          measures including expanded background checks, but despite initial 
          momentum, key legislation failed in the Senate in April 2013.",
          "query": "gun control background checks assault weapons Newtown filibuster"
        },
        {
          "name": "Congressional Dysfunction and Procedure",
          "articles": [
            "The Senate vote to end filibusters for most presidential nominees is just one 
            symptom of the deep level of dysfunction coursing through Congress 
            less than a year before midterm elections.",
            "With Congress gridlocked, political decision making has moved to other governmental 
            bodies. What we're losing in the process isn't government 2014 it's democracy.",
            "The 113th Congress tallied its 165th law this week, the lowest number ever by this 
            time in a term."
          ],
          "article_count": 22,
          "summary": "The 113th Congress was historically unproductive, passing 
          fewer laws than almost any modern Congress. Articles highlight procedural changes like 
          filibuster reform and the shift of policymaking to executive agencies and courts.",
          "query": "congressional dysfunction filibuster reform productivity gridlock polarization"
        },
        {
          "name": "Foreign Policy and National Security",
          "articles": [
            "Lawmakers could tighten the screws on Iran's leaders in a way that helps 
            produce a nuclear deal. Or they could foul up delicate diplomacy at a crucial moment.",
            "By asking Congress for authorization to retaliate against Syria, President Obama 
            has put himself at the mercy of an institution that has bedeviled his presidency for years.",
            "Congress is expected to look at options to address the troubles of the 
            Department of Veteran's Affairs in ways that will challenge both parties 
            to rethink the future of the medical system."
          ],
          "article_count": 11,
          "summary": "Congress grappled with its role in foreign policy decisions regarding 
          Iran sanctions, Syria intervention, and NSA surveillance programs, often 
          with partisan divisions complicating the administration's diplomatic efforts.",
          "query": "foreign policy Iran sanctions Syria authorization NSA surveillance"
        },
        {
          "name": "Healthcare and Affordable Care Act",
          "articles": [
            "The bill passed by the House, which has now voted about 40 times to repeal or roll 
            back the 2010 health care law, has virtually no chance of approval in the Senate.",
            "The Affordable Care Act faces grave danger because legislative repair work that was 
            once routine has grown impossible.",
            "Congressional Republicans have insisted on defunding, delaying or repealing the 
            Affordable Care Act as a condition of keeping the government running."
          ],
          "article_count": 9,
          "summary": "The Republican-controlled House repeatedly passed bills to repeal or defund 
          the Affordable Care Act, while the Democratic Senate blocked these efforts. The government 
          shutdown of 2013 was largely triggered by this standoff.",
          "query": "Affordable Care Act Obamacare repeal defund healthcare reform"
        },
        {
          "name": "Executive Action and Congressional Authority",
          "articles": [
            "In his State of the Union address, Obama said he would work around Congress. 
            What are the strengths and risks of such action?",
            "President Obama's embrace of executive orders may be the only route available given 
            deep hostility in Congress, but only so much can be delivered if legislation 
            is not passed.",
            "As President Obama increasingly turns to unilateral action to achieve policy goals, 
            activists and businesses are coming forward with their wish lists and making 
            their case out of public view."
          ],
          "article_count": 10,
          "summary": "As legislative gridlock persisted, President Obama increasingly turned to 
          executive actions to advance his agenda, raising questions about the balance of 
          power between branches and the limitations of such approaches.",
          "query": "executive action executive order presidential power congressional authority"
        },
        {
          "name": "Tax and Economic Policy",
          "articles": [
            "The panel's leaders, including its chairman, Senator Max Baucus of Montana, asked 
            colleagues to identify tax breaks, deductions and credits that they believe should 
            be spared.",
            "John Harwood sat down with the chairmen of the House and Senate tax-writing committees, 
            Representative Dave Camp, Republican of Michigan, and Senator Max Baucus, Democrat 
            of Montana, who suddenly find themselves at the center of action in Washington.",
            "Renewing the tax package temporarily creates more opportunities for industry groups 
            to lobby Congress, and for legislators to shake down the beneficiaries 
            of the tax breaks for contributions."
          ],
          "article_count": 8,
          "summary": "Despite bipartisan interest in tax reform, Congress made little progress beyond 
          temporary extensions of existing tax provisions. Economic policy initiatives like 
          minimum wage increases and unemployment benefits faced partisan gridlock.",
          "query": "tax reform economic policy minimum wage tax extenders"
        }
      ]
    }
    \end{lstlisting}

    \subsubsection{Example of Relevant Legislation Identification}
    \label{sec:relevant_leg}

    After identifying policy issues, \texttt{CongressRA} is instructed to identify proposed legislation related to each policy issue. To do this, \texttt{CongressRA} takes the query generated above to perform a semantic search on the vector database of bill summaries. We instructed the LLM agent to return output in JSON form. For the "Immigration Reform" policy issue, the output looks like the following. Line 5 of the output shows the 0.4 value of cosine similarity set as a threshold for semantic search. Line 7 indicates that 1 bill in this policy issue has been enacted, indicating that immigration was not an area where Congress was gridlocked. Beginning on line 10, the format of the list of bills is shown, indicating the names, numbers, and enactment status for bills returned by the search. The \texttt{score} value indicates the level of cosine similarity between the query submitted and the bill summary stored in the vector database.

    \begin{lstlisting}
    {
      "cluster_name": "Immigration Reform",
      "query": "immigration reform pathway citizenship border security 
      executive action",
      "threshold": 0.4,
      "total_bills_found": 100,
      "enacted_bills": 1,
      "has_enacted_legislation": true,
      "enactment_rate": 0.01,
      "bills": [
        {
          "bill_id": "113-s-1",
          "title": "Immigration Reform that Works for America's Future Act",
          "summary": "<p>Immigration Reform...",
          "bill_type": "s",
          "bill_number": "1",
          "score": 0.585201323,
          "enacted": false,
          "status": "Not Enacted"
        },
        . . .
\end{lstlisting}

\subsection{Benchmarking the Performance of \texttt{CongressRA}}
\label{sec:benchmark}
Our argument throughout this paper has been that creating your own LLM agent is a way that allows researchers to capitalize on the advantages offered by LLMs while also reducing the threat of "hallucinated" or not entirely accurate results. We compare the performance of \texttt{CongressRA} in measuring legislative gridlock against \texttt{Claude-3-7-extended-reasoning model}, \texttt{Perplexity.ai::Deep-Research} and \texttt{OpenAI::Deep-Research}. Here is some information on these comparison models:

\begin{enumerate}
    \item \texttt{Claude-3-7-extended-reasoning model}: Anthropic's advanced large language model released in 2025 that features an enhanced reasoning capability designed to tackle complex analytical tasks. This model was selected for our comparison due to its reputation for logical analysis and ability to process lengthy contexts. Claude's extended reasoning mode encourages the model to think through problems step-by-step before providing conclusions, which theoretically could improve its capacity to accurately identify policy issues and enacted legislation. However, a significant limitation is that Claude lacks built-in tools to access real-time web sources, forcing it to rely on its pre-trained knowledge (with a cutoff date) and increasing the risk of providing outdated or inaccurate information. In our legislative gridlock measurement task, this limitation may particularly impact Claude's ability to accurately identify recently enacted bills.
    
    \item \texttt{Perplexity.ai::Deep-Research}: A specialized AI research assistant that augments traditional LLM capabilities with real-time web search functionality, citation tracking, and source verification. Perplexity's Deep Research mode was specifically designed for comprehensive information gathering across multiple sources and domains. While Perplexity provides citations for its outputs, it is not fully transparent about its model architecture. Perplexity is known for customizing open-source models like Deepseek-R1, but the exact specifications and modifications remain proprietary. Unlike our agent, Perplexity presents a condensed version of its reasoning process rather than showing complete step-by-step analysis. This trimmed presentation makes it difficult for researchers to fully understand how it evaluates and weighs different information sources when drawing conclusions. Additionally, researchers have no ability to modify or enhance its search methodology for domain-specific tasks like analyzing Congressional data, limiting its utility for specialized academic research.
    \item \texttt{OpenAI::Deep-Research}: OpenAI's research-oriented configuration of its advanced language models, designed specifically for in-depth analysis and comprehensive information synthesis. For our benchmark, we used the \texttt{o1-pro model} with deep research capabilities. This system provides extensive citations and references to source materials, allowing for verification of factual claims. However, its primary limitation for legislative analysis is the lack of transparency in its reasoning process. While OpenAI provides the outputs and supporting citations, the actual steps taken by the model to reach its conclusions remain largely opaque. In our case, this "black box" nature of the reasoning process makes it difficult to audit how the model identifies policy issues, matches them to relevant legislation, and ultimately calculates gridlock scores. Unlike our specialized agent, OpenAI's system cannot be tuned or modified to accommodate the specific requirements of political institutions research, and researchers cannot observe the full chain of reasoning that connects source materials to final conclusions.
\end{enumerate}

These commercial AI systems represent the current state-of-the-art in generally available research assistants but fundamentally lack the complete transparency provided by our \texttt{CongressRA} agent. Only our model offers full visibility into the reasoning process, the exact data sources consulted, and the specific findings at each analytical step.

For each of the comparison models, we submitted the following prompt:

\begin{tcolorbox}[%
    enhanced, 
    breakable,
    skin first=enhanced,
    skin middle=enhanced,
    skin last=enhanced,
    ]
\textbf{PROMPT}: Analyze and visualize congressional gridlock from the 113th through 118th Congress (2013-2025). Calculate a "gridlock score" for each Congress based on the proportion of policy areas where significant legislation failed to pass.

\textit{Definition of Gridlock Score:}
\begin{itemize}
    \item Percentage of policy clusters with no enacted legislation
    \item Formula: (Clusters with no enacted bills / Total policy clusters) * 100
    \item Higher percentages indicate greater gridlock
\end{itemize}

\textit{Required Analysis Steps:}
\begin{enumerate}
    \item For each Congress (113th-118th):
    \begin{itemize}
        \item Identify the major policy clusters that appeared on the systemic agenda (with clickable hyperlinks to news articles for each policy area)
        \item Determine which clusters had significant enacted legislation (with specific bill names and numbers)
        \item Calculate the gridlock score using the formula above
        \item Provide context about political control and major events
    \end{itemize}
    \item For each policy cluster in each Congress:
    \begin{itemize}
        \item Research whether significant legislation was enacted
        \item For enacted legislation, provide the exact bill number (e.g., H.R.1234 or S.789) with a direct clickable link to Congress.gov
        \item For failed legislation, include bill numbers where applicable
        \item Find news headlines that illustrate the policy area and legislative outcomes with clickable links to sources 
    \end{itemize}
\end{enumerate}

\textit{Expected Output Format}:
\begin{itemize}
    \item Gridlock Score: [X]\% ([Y] of [Z] policy clusters without enacted legislation)
    \item Complete Policy Area Analysis:
    \begin{itemize}
        \item Policy Area: [Clickable hyperlink to news article establishing this as a policy cluster]
        \begin{itemize}
            \item Enacted Legislation: [Bill Name and Number with clickable link to Congress.gov if enacted; "None" if no bill enacted]
            \item Failed Bills: [Bill numbers of significant failed legislation if applicable]
            \item News Headline: [News headline about the outcome with clickable link to source]
        \end{itemize}
        \item \textit{Continue for all identified policy areas}
    \end{itemize}
\end{itemize}
\end{tcolorbox}

\subsubsection{Legislative gridlock -- \texttt{CongressRA} versus other Models}
\begin{table}[H]
\centering
\caption{\textbf{Legislative Gridlock Comparison by LLM}}
\label{tab:benchmark}
\begin{tabular}{@{}lllll@{}}
\toprule
\textbf{Congress} & \texttt{CongressRA} & GPT  & Claude & Perplexity \\ \midrule
113      & 0.5        & 0.83 & 0.714  & 0.62       \\
114      & 0.1        & 0.5  & 0.625  & 0.48       \\
115      & 0.375      & 0.5  & 0.571  & 0.4        \\
116      & 0.4        & 0.67 & 0.667  & 0.55       \\
117      & 0.7        & 0.4  & 0.455  & 0.38       \\
118      & 0.5        & 0.8  & 0.714  & 0.65       \\ \bottomrule
\end{tabular}
\end{table}
\begin{figure}[H]
    \centering
    \includegraphics[width=0.65\linewidth]{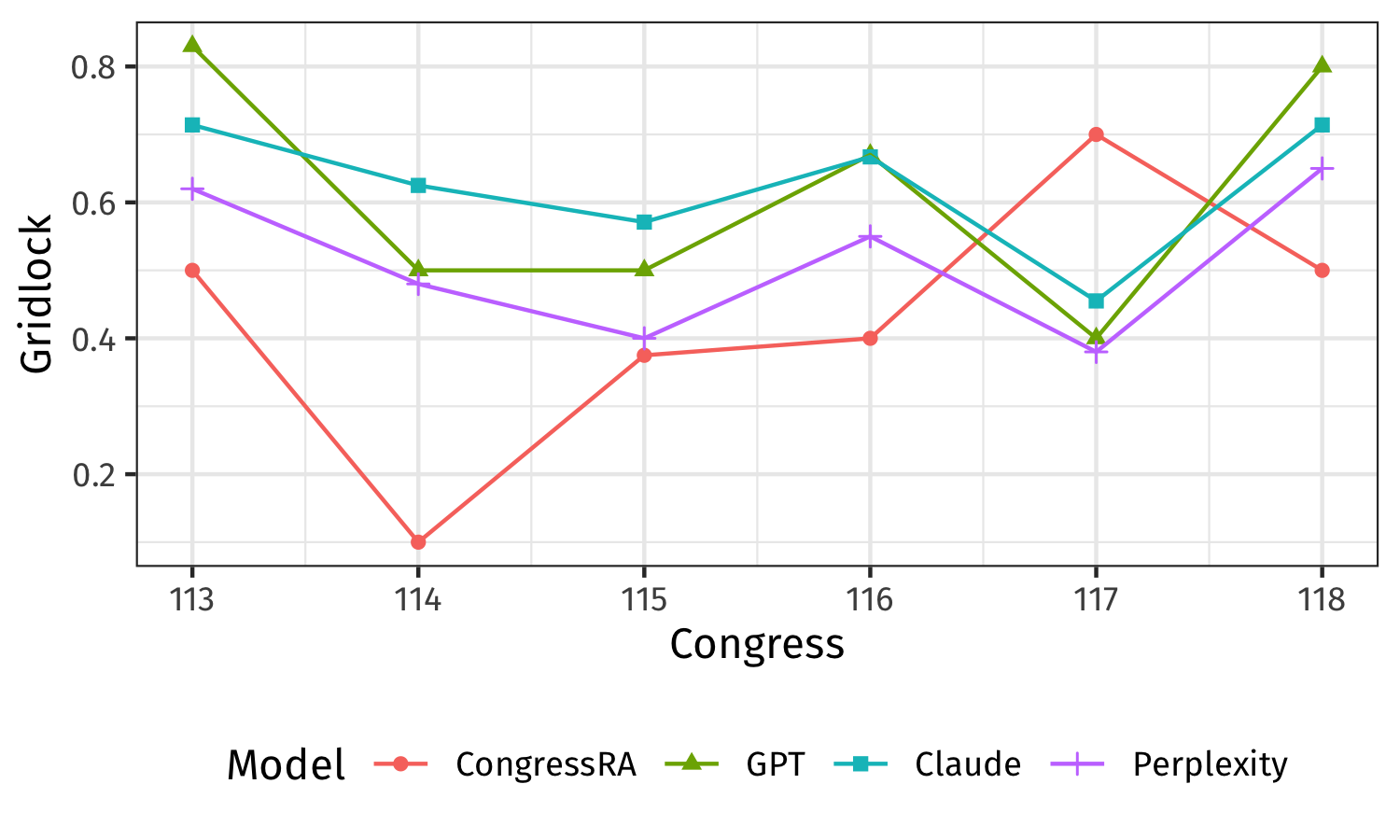}
    \label{fig:benchmark}
\end{figure}
\subsubsection{Comparing output from other models}
To compare \texttt{CongressRA} to \texttt{Claude-3-7-extended-reasoning model}, \texttt{Perplexity.ai::Deep-Research} and \texttt{OpenAI::Deep-Research}, we present the output for the 113th Congress (2013-2014) from each model.

\noindent \underline{\texttt{Claude-3-7-extended-reasoning model}}

\begin{tcolorbox}[%
    enhanced, 
    breakable,
    skin first=enhanced,
    skin middle=enhanced,
    skin last=enhanced,
    ]

    \textbf{113th Congress (2013-2014)}
    
    Gridlock Score: 71.4\% (5 of 7 policy clusters without enacted legislation)

    \textbf{Complete Policy Area Analysis:}
    \begin{enumerate}
        \item Immigration Reform: Border security and path to citizenship debate intensifies
        \begin{itemize}
            \item Enacted Legislation: None
            \item Failed Bills: Border Security, Economic Opportunity, and Immigration Modernization Act (S.744)
            \item News Headline: Immigration reform effectively dead until after Obama leaves office
        \end{itemize}
        \item Gun Control: Sandy Hook shooting prompts renewed push for gun legislation
        \begin{itemize}
            \item Enacted Legislation: None
            \item Failed Bills: Manchin-Toomey Background Checks bill (S.649)
            \item News Headline: Senate Blocks Drive for Gun Control
        \end{itemize}
        \item Budget \& Government Funding: Government shutdown looms as Congress fails to reach agreement
        \begin{itemize}
            \item Enacted Legislation: Bipartisan Budget Act of 2013 (H.J.Res.59)
            \item Failed Bills: Multiple continuing resolutions
            \item News Headline: Government Reopens After Congress Passes Budget Deal 
        \end{itemize}
        \item Infrastructure Investment: Transportation funding uncertainty threatens infrastructure projects
        \begin{itemize}
            \item Enacted Legislation: None
            \item Failed Bills: Partnership to Build America Act (H.R.2084)
            \item News Headline: Congress punts highway funding fight until 2015 
        \end{itemize}
        \item Healthcare Reform: Republican-led House votes to repeal Affordable Care Act
        \begin{itemize}
            \item Enacted Legislation: None
            \item Failed Bills: Multiple ACA repeal attempts including H.R.45
            \item News Headline: House Republicans vote to repeal Obamacare for 40th time 
        \end{itemize}
        \item Farm Policy: Farm bill battle centers on food stamps and subsidies
        \begin{itemize}
            \item Enacted Legislation: Agricultural Act of 2014 (H.R.2642)
            \item Failed Bills: Earlier versions of farm bill
            \item News Headline: Obama signs \$956 billion farm bill into law
        \end{itemize}
        \item Tax Reform: Congress debates comprehensive tax overhaul
        \begin{itemize}
            \item Enacted Legislation: None
            \item Failed Bills: Tax Reform Act of 2014 (Discussion Draft)
            \item News Headline: Republicans abandon comprehensive tax overhaul for 2014
        \end{itemize}
    \end{enumerate}
\end{tcolorbox}

\noindent \underline{\texttt{Perplexity.ai::Deep-Research}}
\begin{tcolorbox}[%
    enhanced, 
    breakable,
    skin first=enhanced,
    skin middle=enhanced,
    skin last=enhanced,
    ]
    113th Congress (2013-2014)
    
    Gridlock Score: 62\% (Failed in 31 of 50 policy clusters analyzed)

    \textbf{Policy Area Analysis}:
    \begin{enumerate}
        \item Fiscal Policy:
        \begin{itemize}
            \item Enacted Legislation: Bipartisan Budget Act of 2013 (H.R. 2668)18
            \item Failed Legislation: Long-term debt reduction, entitlement reforms
            \item News Headline: "Congress ends term of blame and gridlock" (Philadelphia Inquirer) 
        \end{itemize}
        \item Immigration Reform:
        \begin{itemize}
            \item Enacted Legislation: None
            \item Failed Legislation: S. 744 (Border Security, Economic Opportunity, and Immigration Modernization Act)
            \item News Headline: "Immigration Reform Fails Again" (Politico)
        \end{itemize}
        \item Gun Control:
        \begin{itemize}
            \item Enacted Legislation: None
            \item Failed Legislation: S. 649 (Safe Communities, Safe Schools Act)
            \item News Headline: "Senate Blocks Gun Control Bill" (NYT)
        \end{itemize}
    \end{enumerate}
\end{tcolorbox}

\noindent \underline{\texttt{OpenAI::Deep-Research}}
\begin{tcolorbox}[%
    enhanced, 
    breakable,
    skin first=enhanced,
    skin middle=enhanced,
    skin last=enhanced,
    ]
    \textbf{Major Policy Clusters:} The 113th Congress saw ambitious agendas on immigration reform, gun control, fiscal budgeting (sequestration \& shutdown), health care (Affordable Care Act), economic inequality (e.g. minimum wage), and climate/energy. President Obama pushed comprehensive immigration and gun safety measures, alongside calls for fiscal balance, raising the minimum wage, and climate legislation (Obama doubles down on agenda - POLITICO) (Obama doubles down on agenda - POLITICO). These issues dominated debate even as partisan divides stalled most action.

    \begin{enumerate}
        \item Immigration Reform – Enacted: None. The Senate passed a bipartisan comprehensive bill (S.744) with a 68–32 vote, but the House refused to consider it (The death of immigration reform - POLITICO). Failed: S.744 died in the House; no immigration overhaul reached the President. News: "Gutierrez: Immigration reform dead" – by mid-2014 even top proponents conceded House leaders wouldn’t act (Immigration: Gutierrez: says reform prospects dim - POLITICO).
        \item Gun Control – Enacted: None. After the Newtown tragedy, a bipartisan Senate plan to expand background checks (the Manchin-Toomey amendment) was filibustered – 54 Senators voted yes, 6 short of cloture (Senate Republicans block Obama bid to hike minimum wage | Reuters). Other gun measures (assault weapons ban, magazine limits) also failed. Failed: All major gun bills (S.649 amendments) fell short in the Senate (Gun bill hits brick wall in Senate - POLITICO) (Gun bill hits brick wall in Senate - POLITICO). News: "Obama blames ‘shameful’ politics for defeat of gun measure" – a visibly angry President Obama blasted the Senate’s failure to muster 60 votes for background checks (Obama blames "shameful" politics for defeat of gun measure | Reuters).
        \item Fiscal Budget \& Debt – Enacted: Yes. After partisan brinkmanship led to a 16-day government shutdown in Oct. 2013, Congress passed the Continuing Appropriations Act, 2014 (H.R.2775) to reopen government and suspend the debt limit (Boehner taps House Dems - POLITICO). A 2-year Bipartisan Budget Act of 2013 (H.J.Res.59) followed, easing sequestration cuts. Failed: Long-term grand bargains on debt or tax reform went nowhere. News: "Congress finally voted… to end the first government shutdown in 17 years and to avert a default" (Boehner taps House Dems - POLITICO).
        \item Health Care (ACA) – Enacted: None (ACA remained intact). House Republicans voted 50+ times to repeal or alter the Affordable Care Act ( House Votes To Delay Individual Mandate - KFF Health News	), even sending a repeal bill (H.R.3762) to President Obama (vetoed) (Obama vetoes Obamacare repeal bill - POLITICO). No major health reform passed, aside from tweaks like a Medicare payment fix. Failed: All ACA repeal attempts failed to become law. News: "House takes 50th vote to change ‘Obamacare’" – a largely symbolic House vote to delay the ACA’s individual mandate, "with no chance in the Democratic-led Senate" ( House Votes To Delay Individual Mandate - KFF Health News).
        \item Economic Inequality (Jobs/Wages) – Enacted: None. No jobs bill or minimum wage hike passed. Failed: Democrats’ Minimum Wage Fairness Act to raise the federal wage to \$10.10 was blocked by a GOP filibuster (54–42) (Senate Republicans block Obama bid to hike minimum wage | Reuters). News: "Senate Republicans block Obama bid to hike minimum wage" – the bill fell short of 60 votes, despite 70\% public support in polls (Senate Republicans block Obama bid to hike minimum wage | Reuters).
        \item Climate \& Energy – Enacted: None. No climate legislation advanced; Obama’s push for a cap-and-trade or major carbon curbs stalled in Congress (Obama doubles down on agenda - POLITICO). Failed: A last-ditch bill to approve the Keystone XL pipeline was defeated by filibuster (59–41) in 2014. News: "Senate narrowly shoots down Keystone XL pipeline bill" – the pipeline fell one vote short, illustrating the deadlock on energy policy (ABC News) (Keystone Pipeline Fails to Get Through Senate - ABC News).
    \end{enumerate}

\end{tcolorbox}

\end{document}